\documentclass[sigconf]{acmart}

\AtBeginDocument{%
  \providecommand\BibTeX{{%
    \normalfont B\kern-0.5em{\scshape i\kern-0.25em b}\kern-0.8em\TeX}}}

\copyrightyear{2024} 
\acmYear{2024} 
\setcopyright{acmlicensed}\acmConference[HRI '24 Companion]{Companion of the 2024 ACM/IEEE International Conference on Human-Robot Interaction}{March 11--14, 2024}{Boulder, CO, USA}
\acmBooktitle{Companion of the 2024 ACM/IEEE International Conference on Human-Robot Interaction (HRI '24 Companion), March 11--14, 2024, Boulder, CO, USA}
\acmDOI{10.1145/3610978.3640674}
\acmISBN{979-8-4007-0323-2/24/03}

\begin{document}

\title{Learning to Control an Android Robot Head for Facial Animation}


\author{Marcel Heisler}
\affiliation{%
  \institution{Hochschule der Medien}
  \city{Stuttgart}
  \country{Germany}
}
\email{heisler@hdm-stuttgart.de}

\author{Christian Becker-Asano}
\affiliation{%
  \institution{Hochschule der Medien}
  \city{Stuttgart}
  \country{Germany}
}
\email{becker-asano@hdm-stuttgart.de}


\begin{abstract}
The ability to display rich facial expressions is crucial for human-like robotic heads. While manually defining such expressions is intricate, there already exist approaches to automatically learn them. In this work one such approach is applied to evaluate and control a robot head different from the one in the original study. 
To improve the mapping of facial expressions from human actors onto a robot head, it is proposed to use 3D landmarks and their pairwise distances as input to the learning algorithm instead of the previously used facial action units. 
Participants of an online survey preferred mappings from our proposed approach in most cases, though there are still further improvements required.
\end{abstract}

\ccsdesc[100]{Human-centered computing~User studies}
\ccsdesc[500]{Computing methodologies~Robotic planning}

\keywords{android robot, artificial intelligence, facial expression}

\maketitle

\section{Introduction}
An increasing amount of actuators built into robotic faces, enhances such faces' capabilities to convey information using facial expressions. At the same time, the complexity of controlling these facial expressions using manually defined actuator movements increases.

Thus, several methods that learn in an automated way how to control a robot's actuators to generate facial expressions have been proposed \cite{wu_learning_2009, chen_smile_2021, yang_optimizing_2022, rawal_exgennet_2022, auflem_facing_2022}.
Two of these approaches are limited to generating facial expressions only for a set of five \cite{rawal_exgennet_2022} or seven \cite{yang_optimizing_2022} emotions, which is difficult to combine with other facial movements, like speech or eye-gaze, at the same time.
The other approaches learn how to map facial landmarks \cite{chen_smile_2021} or action units (AUs) \cite{wu_learning_2009, auflem_facing_2022} to control commands of robotic faces. These intermediate representations are then used to map facial expressions from human faces onto robots. In the case of AUs, the Facial Action Coding System (FACS) \cite{ekman_facial_1978} can be used to combine AUs into basic emotions.

In addition to extracting these intermediate representations from real humans, it is possible to generate them automatically based on input from other modalities, like speech. E.g., there are machine learning (ML) based approaches that map input speech signals to facial landmarks \cite{eskimez_noise-resilient_2020, bigioi_pose-aware_2022} or facial AUs \cite{meng_listen_2017}. Combining such a mapping from speech to an intermediate representation with a mapping from the intermediate representation to the actuator controls is expected to enable lip-sync speech animations for robot heads based only on automatic ML. This eases the implementation of current approaches and the application to different robotic heads \cite{heisler_making_2023}.

To the best of our knowledge, none of the approaches that learn a mapping from intermediate representations to a robotic head has been applied to and validated on a different robot head than the one in corresponding original works. Though \cite{wu_learning_2009} and \cite{auflem_facing_2022} are similar approaches and achieved promising results for their particular heads used in each study. Additionally, in \cite{auflem_facing_2022} it is highlighted that the approach is not only usable to control the robot head's actuators but also to evaluate its hardware capabilities in a more general way.

Thus, in this paper a reimplementation of \cite{auflem_facing_2022} and its application to a different robot head is presented. After manual inspection of the results some improvements are proposed and compared to our reimplemenation of the original approach using an online survey.

\section{Background}

\subsection{Robot Head}
\label{head_description}
\begin{figure}[htbp]
\centerline{\includegraphics[width=0.35\textwidth]{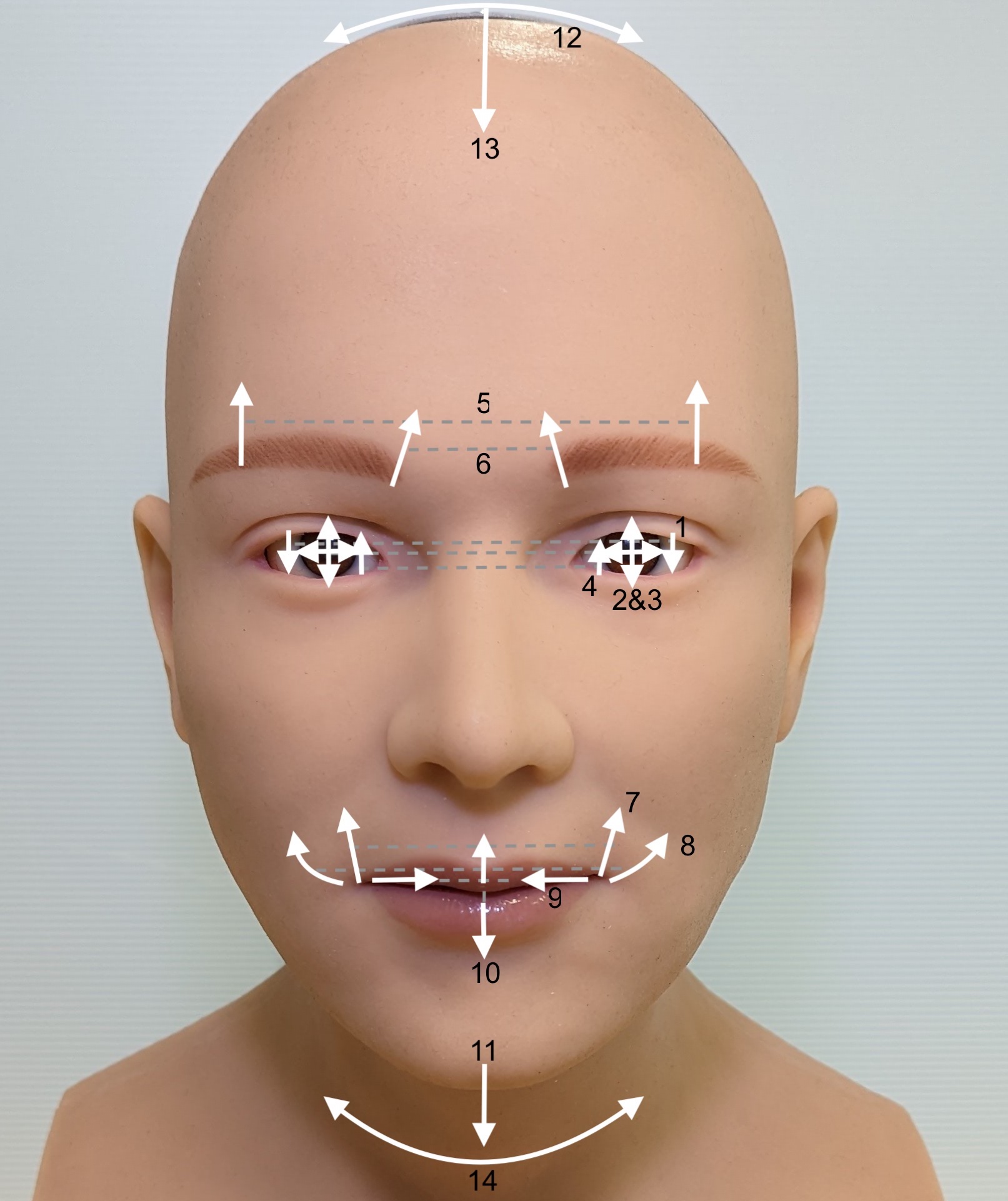}}
\caption{Possible movements of the android robot head. 1: upper eyelid down, 2: eyeball left right, 3: eyeball up down, 4: lower eyelid up, 5: eyebrow up, 6: eyebrow shrink, 7: mouth corner up, 8: mouth corner back, 9: lip shrink, 10: lips open, 11: jaw down, 12: lean head, 13: nod, 14: tilt head. Dotted lines indicate symmetric movements by a single actuator.}
\label{actuators}
\end{figure}

The robot head used in this work is an android robot head built by the Japanese company A-Lab\footnote{https://www.a-lab-japan.co.jp/en/}, as initially described in \cite{heisler_making_2023}. It has 14 pneumatic actuators, that are controlled by sending values in the range of $0-255$ via an RS485 connection. Its possible movements are depicted in Fig.~\ref{actuators}.

\subsection{Facing the FACS}
The work 
by Auflem et.~al \cite{auflem_facing_2022}, that is transferred to our robot head, is briefly summarized in this subsection.

The overall approach consisted of a robot head prototype controlled to do random movements and being recorded by a camera. This recording was analyzed using OpenFace 2.0 \cite{baltrusaitis_openface_2018} to extract the activations of AUs, with values ranging from $0$ to $5$, where $0$ means the AU is not active at all and $5$ indicates the AU is maximally active. 
The extracted AUs together with the random control commands resulted in a dataset that was used to train a learning algorithm to predict motor control commands for facial AUs. 

The robot head reported on in \cite{auflem_facing_2022} has actuation points at the root of the nose, the mouth corner (cheek) and the eyebrows that are actuated using six servo motors. It can also open and close its eyelids. For the creation of the dataset the nose, eyebrow and cheek servos were moved symmetrically and the eyelids were kept open, which lead to three values that needed to be predicted.

As input to the learning algorithms 17 AUs were used. Due to fluctuations in the AU predictions of static facial expressions, the average AU values of seven consecutive frames showing the same expression were used in \cite{auflem_facing_2022}.
To use the person specific normalization implemented in OpenFace2.0 the recording included 75\% neutral facial expressions along with the random ones. 
However, the neutral expressions were not included in the dataset, which finally consists of 500 frames.

To analyze the hardware capabilities Pearson correlation coefficients between all servos and AUs were calculated and inspected. It was found that each AU correlates with at least one of the servos.

To learn the mapping from AUs to servo control values different learning algorithms and hyperparameter settings were compared: linear regression (LR), ridge regression (RR), support vector regression (SVR) and multilayer perceptron (MLP). The MLP achieved the best results regarding the root-mean-square error (RMSE).

For further evaluation FACS was used for maximizing specific AUs to generate facial expressions for six basic emotions (anger, disgust, fear, happy, sadness, surprise). AUs not involved in the corresponding emotion were set to zero. Images of the resulting facial expressions were then classified by Residual Masking Network (RMN) \cite{pham_facial_2021}, which was at this time a state-of-the-art ML model for facial emotion recognition. Only \textit{surprise} and \textit{happy} could be classified correctly.

Finally, there were qualitative results shown from a real-time reenactment application, where AUs were calculated for a human actor's face and mapped to the robot head \cite{auflem_facing_2022}.

\section{(Re-) Implementation}
\begin{figure}[htbp]
\centerline{\includegraphics[width=0.35\textwidth]{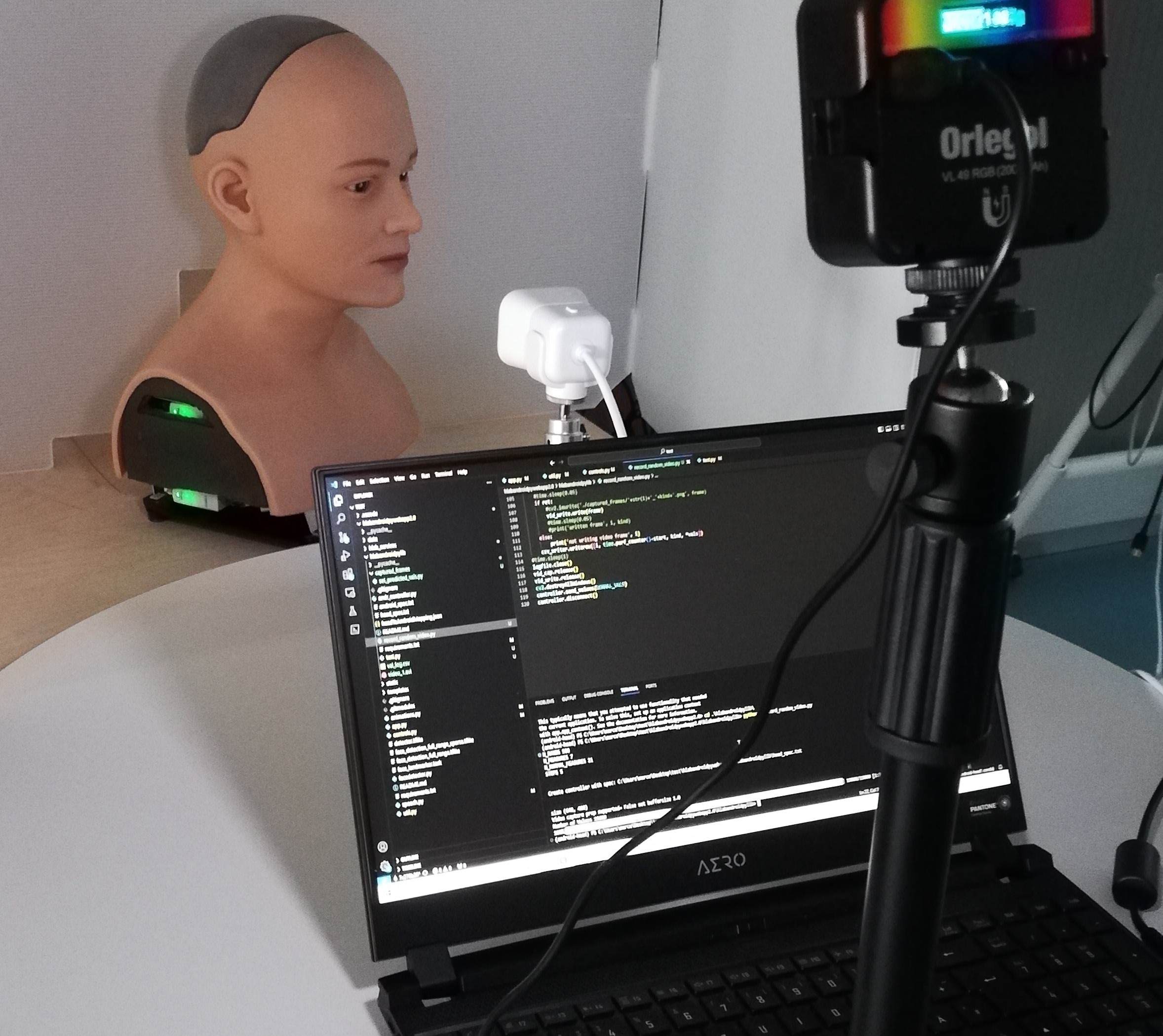}}
\caption{Setup of dataset creation.}
\label{setup}
\end{figure}

In transferring the approach from Auflem et.~al to the robot head described in Section~\ref{head_description} the overall approach is kept the same. As shown in Fig.~\ref{setup} the android robot head is controlled by a notebook to perform random movements. These are recorded using a webcam connected via USB-C. The resulting video is analyzed to extract AUs using OpenFace2.0. An additional LED spotlight is used to establish more stable lighting conditions.

\subsection{Adjustments}
Some adjustments to the original approach were necessary. First, due to availability a different webcam was used, i.e.,~a Logitech StreamCam (960-001297).
Next, between random and normal frames an interpolation over four frames was made to move the facial expression in a smoother way. These interpolation steps are also recorded, but not used as part of the 500 frames for the dataset.  
Sending values to the robot controller and instantly recording the frame using the camera results in a delay between the facial expression actually recorded and the facial expression that should have been recorded. Thus, a small pause of 0.25 seconds after sending the values was made before recording the frame. This value was empirically adjusted until the frames showed the expression as expected on manual inspection.

Only nine of the 14 actuators available in the robot head are used in the dataset. Actuators for eyeball movements (2 and 3) as well as actuators for neck movements (12, 13, 14) are excluded since they should have no impact on AUs and only rarely effect facial expressions.

Only a subset of the learning algorithms from the original work are explored in this work again: LR for its simplicity and MLP because it led to the best results in \cite{auflem_facing_2022}. For the MLP the same set of hyperparameters are tuned using GridSearch, as described in the original work.

For FACS based emotional facial expressions, instead of zero, the respective minima in the train dataset are used as AU values for inactive AUs.

\subsection{Extensions}
Aiming to improve mappings from human actors onto the robot head, the work by Auflem et al.~was extended as described next. 

First, because the robot head is not as expressive as a human, the AU values detected for the human actor are MinMax-scaled to match the distribution of the robot head's values, before motor values are predicted.

Second, representations of facial expressions other than AUs are used as input to the learning algorithm. 
The 68 facial landmarks detected by OpenFace2.0 in 3D space are tested first using an LR model with a dimensionality reduction to 17 dimensions by a Principal Component Analysis (PCA) as preprocessing step. Since this already outperforms the MLP trained on AUs with finetuned hyperparameters, cf.~Table~\ref{rmse}, further experiments with landmarks are conducted.
To map facial expressions from a human face to the robot, the landmarks need to be aligned to the robot's landmarks. To achieve this the landmarks are centered and rotated using the \textit{pose} values predicted by OpenFace2.0, then a Procrustes analysis, as suggested in \cite{eskimez_noise-resilient_2020}, is performed to further align the landmarks.
Finally, pairwise distances between all 3D landmarks are used as input to a LR model to predict actuator controls, since they are expected to be easier align-able from a human actor using MinMax-Scaling. Again, a PCA is used for preprocessing. For the new dimensionality different values are tested: every second value in a range from $3$ to $40$. Twenty-seven dimensions, explaining 99.92\% of the variance in the data, are chosen since higher values hardly further reduced the RMSE on the test set.

\subsection{Results}
\begin{figure}[htbp]
\centerline{\includegraphics[width=0.45\textwidth]{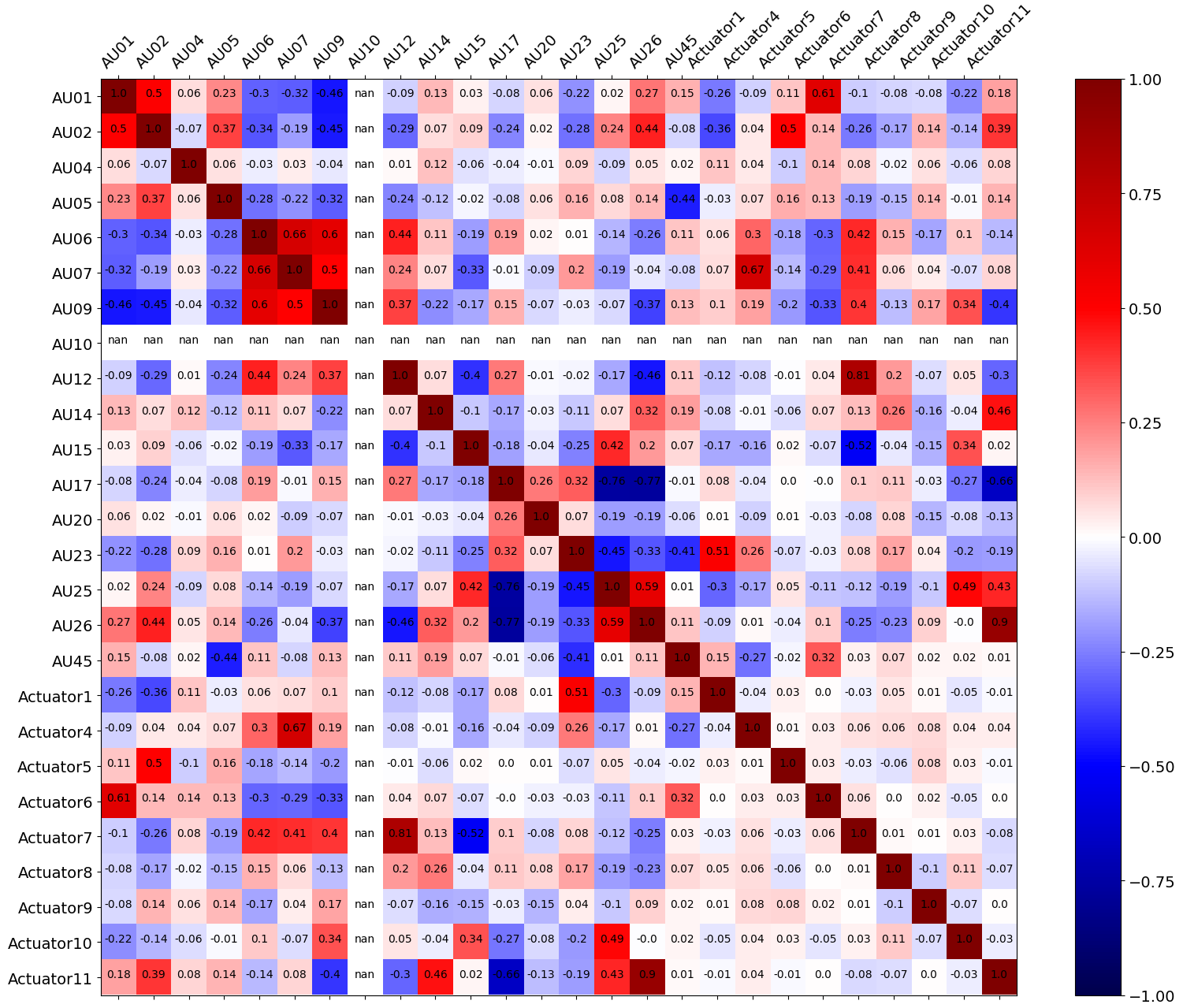}}
\caption{Correlation matrix showing the Pearson correlation coefficients between AUs and actuators.}
\label{correlation_matrix}
\end{figure}

As Auflem et.~al point out a correlation matrix as shown in Fig.~\ref{correlation_matrix} between actuators and AUs can inform the hardware design about possible future improvements. E.g., the absence of any correlations for AU10 highlights the lack of an actuator controlling the roots of the nose of our robot head. Due to this lack of correlations AU10 is not considered for training any learning algorithms. High correlations between AUs are undesirable, since this means that they cannot be controlled separately. They are especially observable for AUs 1 and 2 as well as between 6, 7, and 9. However the correlations between AUs 1 and 2 are much smaller for the robot head used in this work than for the robot head used by Auflem et.~al, which indicates that a second actuator for the eyebrows already helps to disentangle the activation of those AUs. AUs 6, 7, and 9 all correlate with actuator 7, which indicates that the desired movement of moving the mouth corners up also moves up the cheeks and even effects the tightening of the eyelids and a wrinkling of the nose. Most actuators correlate in expected ways with AUs, e.g., the jaw actuator (11) has the strongest correlation with AU26. However, there are also surprising correlations to be found like actuator 1 (moving the upper eyelid) that correlates stronger with AUs 1 and 2 (eyebrows) or even AUs 23 and 25 (lip movements) than with AU45 (blink). Such correlations require further investigation.

\begin{table*}[htbp]
    \centering
    \begin{tabular}{lcccccc} \hline 
         \textbf{Expression}&  \textbf{Anger}&  \textbf{Disgust}&  \textbf{Fear}&  \textbf{Happy}&  \textbf{Sadness}& \textbf{Surprise}\\ \hline 
         Maximized AUs&  4, 7, 23&  9, 15&  1, 2, 4, 5, 7, 20, 26&  6, 12&  1, 4, 15& 1, 2, 5, 26\\ \hline 
         Result& \includegraphics[scale=0.25]{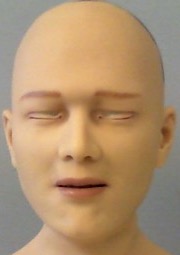} & \includegraphics[scale=0.25]{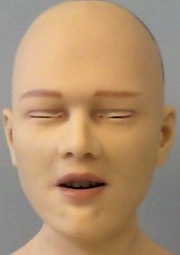} & \includegraphics[scale=0.25]{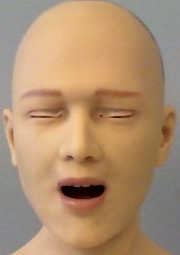} & \includegraphics[scale=0.25]{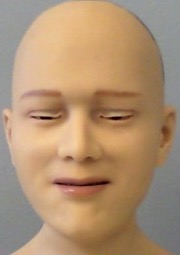} & \includegraphics[scale=0.25]{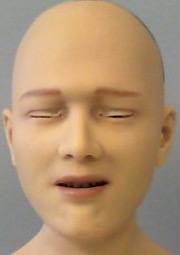} & \includegraphics[scale=0.25]{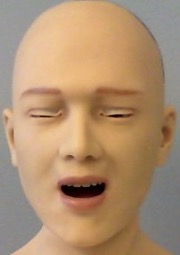} \\ \hline 
         RMN predictions&  neutral: 0.59&  surprise: 0.38&  sad: 0.46&  neutral: 0.45&  sad: 0.53& sad: 0.45\\ \hline
    \end{tabular}
    \caption{FACS based generated facial expressions with top RMN predictions and confidence.}
    \label{emos}
\end{table*}

Comparing different learning algorithms on the test set of AUs to predict actuator control values shows--similar to the results of Auflem et.~al--that an MLP with finetuned hyperparameters outperforms an LR model at least slightly for each actuator, cf.~Table~\ref{rmse}. When the RMSEs reported by Auflem et.~al are rescaled to the value range used here ($0-255$) these values are much better than the results achieved here (Cheek servo: 10.41, Eyebrow servo: 17.15 and Nose servo: 15.09). 

Worse predictions on the test set suggest, that also the creation of emotional facial expressions using FACS becomes less feasible. As can be seen in Table~\ref{emos} this is the case and RMN correctly predicted the intended emotion only for \textit{sad}.

Due to the unexpected correlations found in Fig.~\ref{correlation_matrix}, it is expected that the poor results can partially arise from an imprecise detection of AUs. Thus, facial landmarks are used as input to an LR model. Without any parameter tuning, much better results on the test set are achieved, cf.~Table~\ref{rmse}, third column. Instead of the raw landmarks the pairwise distances between them are used, since they appear easier to rescale from human faces to the robot's values. The fourth column of Table~\ref{rmse} shows their results on the test set. In contrast to the landmarks, the number of dimensions reduced to using PCA is optimized as a hyperparameter. This results in the best RMSEs on the testset for each actuator.

\begin{table}[htbp]
    \centering
    \begin{tabular}{|c|cccc|} \hline 
         \textbf{Act.}&  \textbf{AUs + LR}&  \textbf{AUs + MLP}&  \textbf{Landm. + LR}& \textbf{Dist. + LR}\\ \hline 
         \textbf{1}&  43.04&  39.74&  23.66& 20.46\\
         \textbf{4}&  49.68&  44.40&  34.05& 33.14\\
         \textbf{5}&  58.47&  57.10&  27.47& 21.73\\
         \textbf{6}&  49.26&  45.12&  29.53& 23.99\\
         \textbf{7}&  38.65&  35.44&  36.91& 29.94\\
         \textbf{8}&  64.04&  63.11&  50.72& 39.87\\
         \textbf{9}&  67.56&  65.31&  42.58& 29.30\\
         \textbf{10}&  59.70&  55.84&  41.06& 33.45\\
         \textbf{11}&  22.54&  22.49&  9.90& 9.26\\
         \hline 
    \end{tabular}
    \caption{RMSEs per actuator on test set for different kinds of input data + learning algorithm combinations.}
    \label{rmse}
\end{table}

\begin{table}[htbp]
\centering
\begin{tabular}{|c|c|c|c|}
\hline
\textbf{No.} & \textbf{Human} & \textbf{AUs}& \textbf{Dists} \\
\hline
\textbf{1} & \includegraphics[scale=0.4]{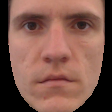} & \includegraphics[scale=0.15]{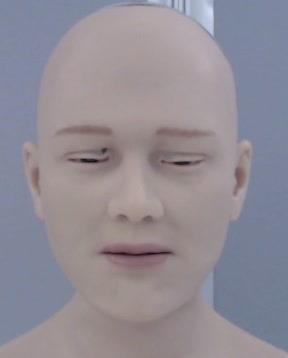} & \includegraphics[scale=0.15]{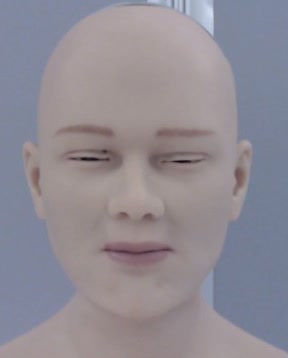} \\
\hline
\textbf{2} & \includegraphics[scale=0.4]{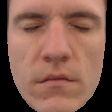} & \includegraphics[scale=0.15]{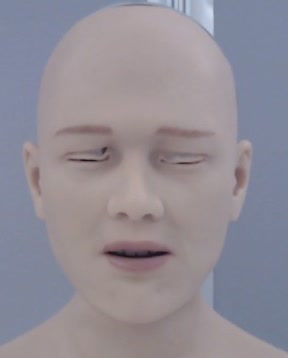} & \includegraphics[scale=0.15]{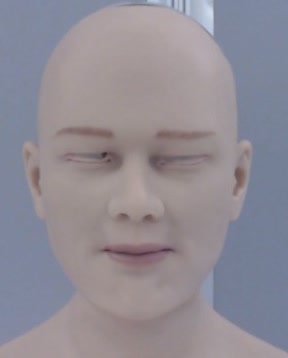} \\
\hline
\textbf{3} & \includegraphics[scale=0.4]{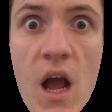} & \includegraphics[scale=0.15]{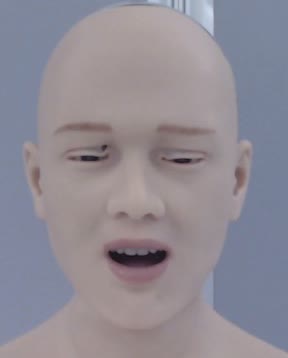} & \includegraphics[scale=0.15]{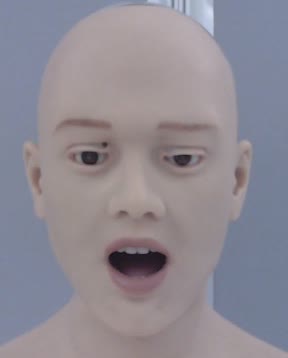} \\
\hline
\textbf{4} & \includegraphics[scale=0.4]{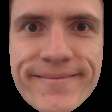} & \includegraphics[scale=0.15]{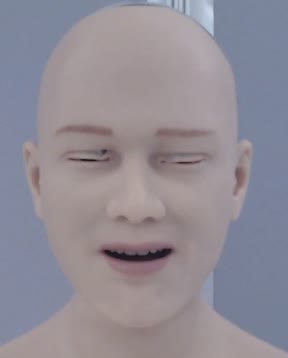} & \includegraphics[scale=0.15]{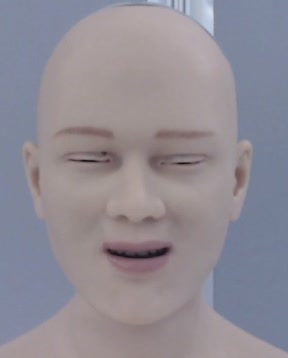} \\
\hline
\textbf{5} & \includegraphics[scale=0.4]{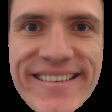} & \includegraphics[scale=0.15]{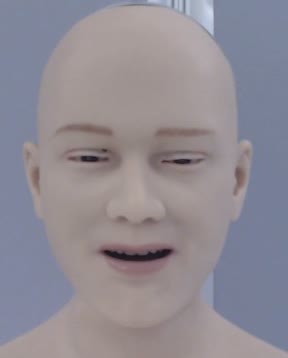} & \includegraphics[scale=0.15]{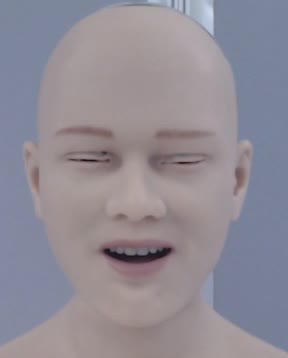} \\
\hline
\textbf{6} & \includegraphics[scale=0.4]{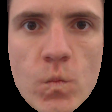} & \includegraphics[scale=0.15]{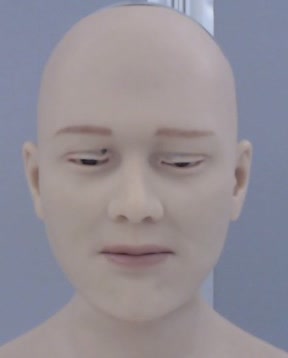} & \includegraphics[scale=0.15]{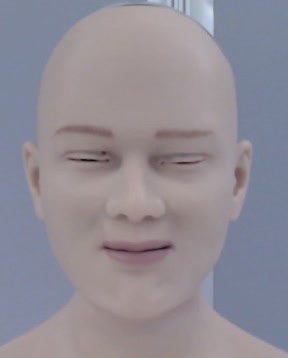} \\
\hline
\end{tabular}
\caption{Qualitative samples for human mapping}
\label{mappings}
\end{table}

\section{Evaluation}
Table~\ref{mappings} shows examples of mappings from a human face to the robot head using AUs as well as pairwise distances. Using these examples, an online survey was conducted. Since these are still preliminary results, the survey was kept simple by asking participants to vote which one of the two mappings they find to look more similar to the image of the human following a forced-choice design.

\begin{figure}[htbp]
\centerline{\includegraphics[width=0.35\textwidth]{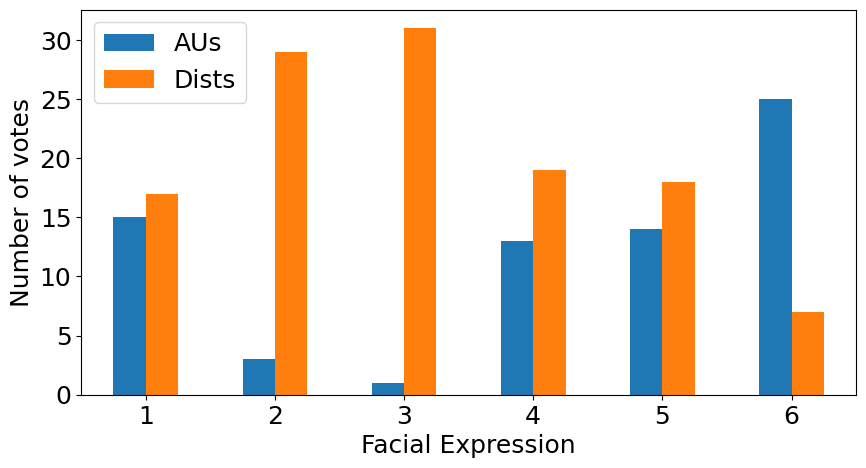}}
\caption{Survey results (cf.~Table~\ref{mappings} for facial expressions)}
\label{survey}
\end{figure}

Thirty-two participants took part in the survey (21 male and 11 female) all aged between 19 and 36 (25 on average). The results are shown in Fig.~\ref{survey}. 
The wide opened mouth and eyes in image 3 look better using the distance-based mapping. This mapping is also clearly favored by the participants for image 2, most likely because with the AU mapping the robot's mouth is not entirely shut. For most other images the votings from the participants are less clear, which makes sense since both mappings are not that good, mainly because the head's eyes are often closed although the human's eyes are not. For image 6 the result in favor of the AU mapping is more surprising since actuator 9 (lip shrink) is actuated stronger using the distance-based mapping.

\section{Conclusion}
In this paper a previous approach to learn a mapping between facial AUs and the actuator controls of an android robot head was applied to a different robot head. Facial landmarks and their pairwise distances are proposed to be used as alternatives to AUs. Using these representations as input to predict actuator controls, performed better on a held out testset of data, than using AUs. An online survey shows promising results for the application of the proposed methods. However, further investigations are needed to align these representations of facial expressions between humans and robot heads. Thus, in future work we plan to try out different scaling and alignment methods as proposed in \cite{chen_smile_2021, eskimez_noise-resilient_2020}. An improved version of the approach described here would simplify the control of different android robot heads and ease their application to real world use-cases.

\bibliographystyle{ACM-Reference-Format}
\bibliography{references}

\end{document}